**Quantile–Physics Hybrid Framework for Safe-Speed Recommendation under Diverse Weather Conditions Leveraging Connected Vehicle and Road Weather Information Systems Data**

**Wen Zhang**
Department of Computer Science and Engineering
University at Buffalo, Buffalo, NY 14260
Email: wzhang59@buffalo.edu

**Adel W. Sadek, Ph.D.**
Department of Civil, Structural and Environmental Engineering
University at Buffalo, Buffalo, NY 14260
Email: asadek@buffalo.edu

**Chunming Qiao, Ph.D.**
Department of Computer Science and Engineering
University at Buffalo, Buffalo, NY 14260
Email: qiao@buffalo.edu

Word Count: 4614 words + 6 table (250 words per table) = 6,114 words

Submitted: July 31, 2025

*Zhang, Sadek, and Qiao*

**ABSTRACT**

Inclement weather conditions can significantly impact driver visibility and tire–road surface friction, requiring adjusted safe driving speeds to reduce crash risk. This study proposes a hybrid predictive framework that recommends real-time safe speed intervals for freeway travel under diverse weather conditions. Leveraging high-resolution Connected Vehicle (CV) data and Road Weather Information System (RWIS) data collected in Buffalo, NY, from 2022 to 2023, we construct a spatiotemporally aligned dataset containing over 6.6 million records across 73 days. The core model employs Quantile Regression Forests (QRF) to estimate vehicle speed distributions in 10-minute windows, using 26 input features that capture meteorological, pavement, and temporal conditions. To enforce safety constraints, a physics-based upper speed limit is computed for each interval based on real-time road grip and visibility, ensuring that vehicles can safely stop within their sight distance. The final recommended interval fuses QRF-predicted quantiles with both posted speed limits and the physics-derived upper bound. Experimental results demonstrate strong predictive performance: the QRF model achieves a mean absolute error of 1.55 mph, with 96.43% of median speed predictions within $\pm 5$ mph, a PICP (50%) of 48.55%, and robust generalization across weather types. The model's ability to respond to changing weather conditions and generalize across road segments shows promise for real-world deployment, thereby improving traffic safety and reducing weather-related crashes.

## INTRODUCTION

Traffic safety remains a critical concern worldwide, with excessive speed identified as a major contributing factor to crash frequency and severity. While posted speed limits are widely used to regulate driving speeds, these limits are typically static and fail to account for real-time variations in roadway and environmental conditions such as precipitation, pavement friction, visibility, and traffic flow. Under adverse conditions, the mismatch between fixed limits and actual roadway safety conditions can significantly increase crash risk (*1*). Dynamic safe speed prediction, which adjusts recommended speeds in real time according to prevailing conditions, has therefore gained increasing attention as a promising approach for enhancing roadway safety and operational efficiency (*2*).

The purpose of this study is to develop a dynamic safe speed recommendation framework that integrates data-driven prediction with physical constraints to ensure both accuracy and feasibility. Specifically, we leverage Quantile Regression Forests (QRF) (*3*) to generate adaptive speed intervals that capture the variability of observed driving speeds under different conditions. To guarantee safety and physical interpretability, we further incorporate a physics-based stopping distance constraint to ensure that recommended upper-bound speeds remain achievable within the available sight distance (*4*). The proposed framework is evaluated using Connected Vehicle (CV) data and Road Weather Information System (RWIS) data collected in the Buffalo–Niagara metropolitan area, and its performance is compared against existing approaches.

The paper is organized as follows. First, we review related work on variable speed limits and weather-responsive speed modeling. Next, we introduce the data sources and describe the dataset construction. We then present the proposed methodology, which includes the spatiotemporal processing of freeway data, the modeling approach based on QRF, the physics-based speed limits, and the final speed interval fusion strategy. After that, we report the experimental results, including an evaluation of the QRF model's performance, generalization analysis across varying weather conditions, comparisons with baseline methods, and a representative case study. Finally, we summarize the key findings of the study.

## BACKGROUND

Variable speed limits (VSLs) dynamically adjust posted speeds using real-time information on roadway conditions such as traffic speed, weather, and pavement state. By tailoring speed limits to prevailing conditions, VSLs can improve safety performance and operational efficiency through speed harmonization, reducing speed variance and mitigating the severity of crashes. In addition, VSLs can mitigate the impacts of adverse weather conditions. (*2*) RWIS networks—now maintained by most state DOTs—continuously monitor pavement temperature, surface state, wind, and visibility, providing a critical data backbone for weather-responsive traffic management (*5*). Together, these advancements have laid a strong foundation for predictive models that move beyond static thresholds toward probabilistic, context-aware speed recommendations.

Despite encouraging field results, two technical challenges, which await further research, limit current practice. First, many data-driven models—including neural networks and mean-regression forests—operate as "black boxes," producing point forecasts that lack physical interpretability. Without explicit braking or sight-distance constraints, some recommended speeds may be physically unattainable, while others can be overly conservative. Second, physics-only approaches overlook the rich, nonlinear influence of combined traffic-weather states. Quantile-based models such as QRF can offer calibrated prediction intervals, but they are rarely coupled with mandatory stopping-sight-distance (SSD) constraints. To bridge this gap, our framework integrates QRF-derived percentiles with SSD limits, producing adaptive intervals that are narrow, statistically well-calibrated, and provably safe under the prevailing road–weather envelope.

## DATA

In this study, we use three datasets. They are Connected Vehicle data, RWIS data, and OpenStreetMap (OSM) data. Each dataset is briefly described below.

### Connected vehicle data

CV data comes from Wejo's Vehicle Movements dataset, which captures high-frequency vehicle movement records collected from on-road vehicles. This dataset provides vehicle journey details at intervals of 1 to 5 seconds, including GPS location, instantaneous speed, and other movement-related features. On average, it covers approximately 5–10% of all vehicles on the road, offering a rich sample of real-world driving behavior.

For this study, we obtained data covering the Buffalo-Niagara metropolitan area covers September 28th, 2022 to October 28th, 2022, December 12th, 2022 to December 22nd, 2022, and March 1st, 2023 to March 31st, 2023, a total of 73 days. The data is organized on a daily basis. Each day includes 24 hourly folders, and each hourly folder contains 20 to 40 Parquet files. The total size of each day's data is approximately 6.7 GB. On average, each day includes around 40 million rows, with each row containing 15 attributes. Table 1 summarizes the key attributes in the dataset, along with their units and descriptions.

**TABLE 1 Key Attributes in Vehicle Movements dataset**

| Name | Unit | Description |
|---|---|---|
| dataPointId | string | Unique identifier for each captured data point. |
| journeyId | string | Identifier for a single vehicle journey, from engine ignition on to off. |
| capturedTimestamp | string | Timestamp of when the data point was recorded, including location time offset. |
| latitude | degrees | Captured latitude to 6 decimal places. |
| longitude | degrees | Captured longitude to 6 decimal places. |
| vehicle status ignitionStatus | categorical | String representation of 3 ignition statuses captured as KEY_ON, MID_JOURNEY, KEY_OFF |
| speed | km/h | Speed that the vehicle was travelling at the time datapoint was captured. |

### Road Weather Information System Data

The RWIS data used in this study was collected from a RWIS station powered by Vaisala installed on the University at Buffalo (UB) North Campus in Buffalo, NY. The RWIS serves as a critical roadside sensing unit, continuously monitoring both atmospheric and pavement conditions. It plays a vital role in supporting transportation agencies, particularly during severe weather conditions, by providing real-time insights that improve roadway maintenance and safety operations.

This RWIS station records essential weather and road surface information at 10-minute intervals, resulting in 144 data points per day. Figure 1 presents the distribution of observations categorized by weather group. We obtained data that aligns with the same time periods covered by the connected vehicle dataset: September 28[th], 2022 to October 28[th], 2022, December 12[th], 2022 to December 22[nd], 2022, and March 1[st], 2023 to March 31[st], 2023. Table 2 summarizes the key attributes in the data, along with their units and descriptions.

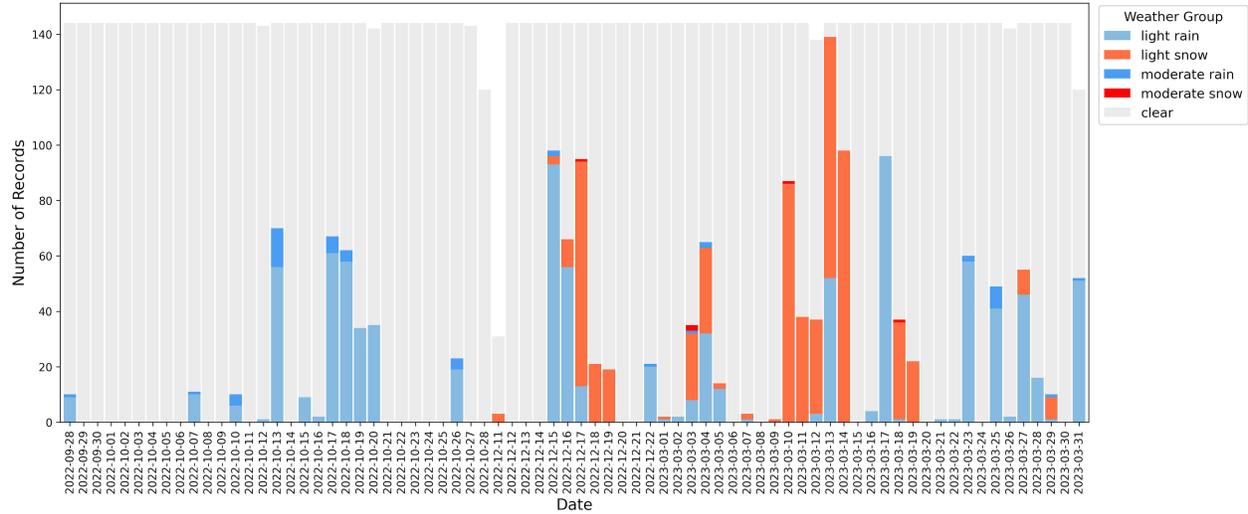

**Figure 1. Stacked bar chart showing the number of RWIS observations per day, grouped by weather condition. Each day includes up to 144 10-minute records. Colors indicate weather categories such as rain, snow, and clear conditions.**

**TABLE 2 Key Attributes in RWIS dataset**

| Name | Unit | Description |
|---|---|---|
| Timestamp | string | Date and time of measurement (every 10 minutes) |
| Surface_tem | Celsius | Road surface temperature |
| Grip | [0, 1] | Surface grip level (0 indicates no grip; 1 indicates maximum grip) |
| Rain State | categorical | Coded weather condition (e.g., light rain, moderate rain, light snow, moderate snow) |
| Visibility | meter | Road visibility (range: 0 − 2000 meters) |
| Precipitation_1 | millimeter | Rolling average precipitation over the past 1 hour |
| Precipitation_3 | millimeter | Rolling average precipitation over the past 3 hour |
| Precipitation_6 | millimeter | Rolling average precipitation over the past 6 hour |
| Precipitation_12 | millimeter | Rolling average precipitation over the past 12 hour |
| Precipitation_24 | millimeter | Rolling average precipitation over the past 24 hour |

**OpenStreetMap data**

We used road network data from OpenStreetMap (OSM) covering the Buffalo, NY region, bounded by the coordinates (N: 42.971123, S: 42.822918, E: −78.692246, W: −78.922892). The road network was extracted using driving mode, which includes all drivable roads accessible to motor vehicles.

The OSM dataset provides rich roadway attributes such as the number of lanes, functional road class (e.g., motorway, primary), posted speed limits, geometric centerlines, and road names. These features are essential for spatially aligning the connected vehicle and RWIS data, enabling accurate map matching and segment-level analysis of traffic and weather conditions. Table 3 summarizes the key attributes in the OpenStreetMap data, along with their units and descriptions.

**TABLE 3 Key Attributes in OpenStreetMap dataset**

| Name | Unit | Description |
|------|------|-------------|
| osmid | string | Unique OSM identifier for the road segment |
| lanes | count | Number of road lines |
| highway | categorical | Road class (e.g., motorway, residential, primary, secondary) |
| maxspeed | mph | Posted speed limit of the segment |
| geometry | geometry | Geographic polyline representing the road centreline |
| name | string | Name of the road |

## METHODOLOGY

### Spatiotemporal Processing of Freeway Data

This study focuses on freeway segments in the Buffalo area with a posted speed limit of 55 mph. Since the connected vehicle data does not include road segment information, we used OSM data to identify and extract vehicle observations located on freeway segments.

Each OSM road segment provides centerline geometry and the number of lanes. According to AASHTO's design standards for the Interstate System (*6*), the standard lane width on the interstate system is 12 feet. We constructed a buffer around each centerline with a radius set to half the total road width to approximate the full road width, see **Equation 1**.

$$\text{Buffer radius} = \frac{\text{number of lanes} \times 12 \text{ ft}}{2}, \tag{1}$$

A spatial join was then performed to retain only connected vehicle records located within these buffered highway areas (Figure 2). It is worth noting that we excluded freeway entrance and exit ramps from the analysis (shown in light green in Figure 2), as driving behavior in these areas does not reflect typical freeway conditions.

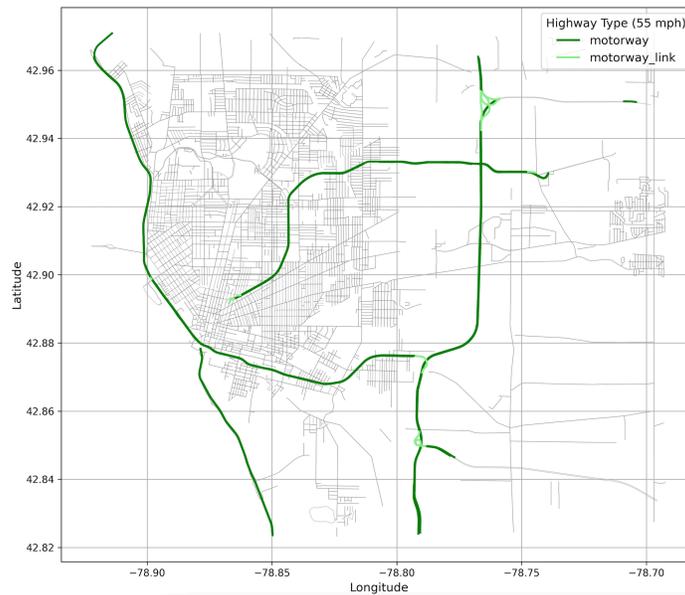

**Figure 2 Road segments with 55 mph speed limit, colored by highway type**

After extracting connected vehicle data on freeway segments, we computed the average speed of each vehicle over 10-minute time windows across the entire study area. For each time window, we then calculated the 0.25 quantile, 0.75 quantile, and median speed across all vehicles.

To integrate weather context, we aligned the connected vehicle data with RWIS data based on timestamp. The resulting dataset contains each vehicle's average speed within 10-minute windows, along with the corresponding weather conditions at that time.

## Modeling Approach

### Quantile Regression Forests

To estimate the conditional speed distribution under varying weather and road conditions, we employ Quantile Regression Forests (QRF) (Meinshausen, 2006) (*3*). Given a target variable $Y$ and predictor variables X, QRF estimates the conditional cumulative distribution function (CDF) of $Y$ for a new input $x$ as:

$$\hat{F}(y|X = x) = \sum_{i=1}^{n} w_i(x) \, 1_{\{Y_i \leq y\}}, \tag{2}$$

where $n$ is the number of training samples. $Y_i$ is the target value (vehicle-level average speed in a 10-minute window) of sample $i$. $y$ is the candidate threshold used to build the CDF. $w_i(x)$ is the weight assigned to training sample $i$, computed as the fraction of trees in which both $x$ and $X_i$ fall into the same leaf node, divided by the size of that leaf. Once $\hat{F}(y|X = x)$ is obtained, the conditional quantile $Q_\alpha(x)$ is defined as:

$$Q_\alpha(x) = \inf \{y: \hat{F}(y|X = x) \geq \alpha\}, \tag{3}$$

which represents the smallest value of $y$ such that the estimated cumulative probability exceeds $\alpha$. This approach enables the estimation of prediction intervals such as $Q_{25} - Q_{75}$ capturing the uncertainty in vehicle speed.

The QRF model is trained on the processed freeway dataset, which contains 26 input features (predictor variables X), including meteorological variables (e.g., surface temperature, visibility, wind speed), pavement conditions (e.g., grip, snow/ice/water surface layers), and contextual features (e.g., hour of day, day of week, vehicle count). The target variable $Y_i$ is the average speed of an individual vehicle in a 10-minute window; a single window may therefore contain multiple $Y_i$ observations. The testing set contains the same features, and the model outputs the 0.25, 0.50, and 0.75 quantile predictions for each 10-minute window.

We use $n_{estimators} = 200$, min_samples_leaf = 10, and the default setting of unrestricted tree depth (max_depth = None). Full parallelization was enabled (n_jobs = -1), utilizing all available CPU cores (12 cores on our system) during training. The total training time is approximately 70 minutes.

### Physics-Based Speed Limits

To ensure safety-aware speed recommendations, we compute a physics-based upper speed limit for each 10-minute observation, denoted as $v_{phys}$. This value represents the highest speed at which a vehicle can come to a complete stop within the driver's visible range under the prevailing road surface conditions. The calculation is as follows.

1. Stopping distance components

The total emergency stopping distance is modeled as:

$$d_{total} = \underbrace{\frac{v^2}{2\mu g}}_{d_{braking}} + \underbrace{v \cdot t_{reaction}}_{d_{reaction}} + \underbrace{k \cdot v}_{d_{gap}}, \tag{4}$$

where: $v$ is the vehicle speed (mph). $\mu$ is the pavement friction coefficient (RWIS grip). $g$ is gravitational acceleration ($32.174 \; ft/s^2$). $t_{reaction}$ is driver reaction time (s). $k$ is an additional headway time for a safety gap (s). The three terms in **Equation 4** correspond to brakeing distance (*7*), reaction distance, and a fixed time-gap buffer.

2. Visibility constraint

To guarantee that the vehicle can safely stop within the distance the driver can see, we define:

$$d_{visible} = \min\left(\text{RWIS}_{visibility}, 495 \text{ ft}\right), \tag{5}$$

where 495 feet corresponds to the AASHTO stopping sight distance for 55 mph (*4*), used as an upper bound to avoid overly permissive limits in exceptionally clear conditions.

3. Closed-form definition of $v_{phys}$

The physics-constrained speed limit is the maximal speed satisfying **Equation 4** with respect to **Equation 5**:

$$v_{phys} = \max\left\{v: \frac{v^2}{2\mu g} + v \cdot t_{reaction} + k \cdot v \leq d_{visible}\right\}, \tag{6}$$

Equation (*6*) is solved numerically for each 10-min record using the corresponding RWIS grip and visibility.

*Interval Fusion*

For deployment, we convert the probabilistic output into a single, operational speed range. Let $Q_{25}$ and $Q_{75}$ be the model-predicted inter-quartile limits and let $v_{phys}$ be the physics-based maximum safe speed under current conditions. We first compute the upper bound.

$$v_{high} = \min\{Q_{75}, v_{phys}, v_{law}\},$$

where $v_{law}$(55mph) is the posted speed limit. The lower bound is

$$v_{low} = \min\{Q_{25}, v_{high}\},$$

and the recommended speed interval for each 10-minute window is

$$\text{Speed interval} = [v_{low}, v_{high}], \tag{7}$$

Since $Q_{25}$ is almost always below $v_{law}$ on the Buffalo network, the lower bound remains effectively data driven; in rare cases where $Q_{25}$ exceeds the legal or physical cap, the interval collapses to a single conservative value ($v_{low} = v_{high}$). The upper bound is capped by both the posted limit and the maximum speed at which a vehicle can stop within the current sight distance and pavement friction. This design keeps the interval centered on observed traffic behaviour, while ensuring that the recommended ceiling never violates legal or the physical safety constraint. The recommended speed interval adapts in real time to road and weather changes while remaining robust to transient traffic fluctuations.

## EXPERIMENTS AND RESULTS

### Experimental Setup
The dataset contains 6,644,562 records and is split into training and testing sets based on timestamps. The training set consists of two continuous periods: September 28[th] to October 18[th], 2022, and December 12[th], 2022, to March 21[st], 2023. The remaining data is used for testing. Both the training and testing sets include a wide range of weather conditions, including rain and snow, ensuring model exposure to diverse environmental scenarios.

### Evaluation Metrics
We adopted the following evaluation metrics to assess the quality of prediction intervals and point estimates ($\delta$):

1. Prediction Interval Coverage Probability (PICP):

$$\text{PICP} = \frac{1}{N}\sum_{i=1}^{N} 1\{y_i \in [l_i, u_i]\}, \tag{8}$$

where $N$ is the number of samples, $y_i$ is the true observation, and $[l_i, u_i]$ is the predicted interval. A higher PICP indicates better coverage of the true observations.

2. Mean Prediction Interval Width (MPIW):

$$\text{MPIW} = \frac{1}{N}\sum_{i=1}^{N} (u_i - l_i), \tag{9}$$

which measures the average width of prediction intervals. Lower MPIW indicates tighter intervals.

3. Mean Absolute Error (MAE):

$$\text{MAE} = \frac{1}{N}\sum_{i=1}^{N} |y_i - \hat{y}_i|, \tag{10}$$

where $y_i$ is the ground truth value, $\hat{y}_i$ is the predicted value (e.g., $Q_{50}$). Lower MAE reflects more accurate central estimates.

4. Threshold Accuracy ($\pm\delta$ mph):

$$\text{Accuracy} = \frac{1}{N}\sum_{i=1}^{N} 1\{|y_i - \hat{y}_i| \le \delta\} \times 100\%, \tag{11}$$

which represents the percentage of time windows in which the absolute prediction error is within a given tolerance $\delta$ (e.g., $\pm 5$mph).

These metrics jointly evaluate the reliability (PICP), sharpness (MPIW), and point accuracy (MAE, RMSE, MAPE, and $\pm \delta$ mph accuracy) of the model.

## Model Performance and Discussion

### *QRF Model Accuracy and Generalization*

The results demonstrate that the model maintains robust generalization from training to testing. At the vehicle level, the 50% PICP decreases only slightly, from 49.65% to 48.55%, remaining very close to the ideal 50% target and indicating that interval calibration holds on unseen vehicles. At the window level, the MPIW rises by 0.24 mph ($\approx 1.3\%$), suggesting the model sensibly widens its intervals under challenging, unseen conditions while still keeping them tight relative to normal freeway speed variance. Although the median-speed MAE increased from 0.73 mph to 1.55 mph, it remains well below the practical 5 mph safety margin; accordingly, 96.43% of 10-minute windows exhibit absolute errors within $\pm 5$ mph. Overall, these results show no evidence of over-fitting and confirm that the model provides well-calibrated prediction intervals and practically useful accuracy for freeway speed estimation.

Table 4 summarizes the training and testing performance of QRF model. Root mean squared error (RMSE, mean absolute percentage error (MAPE) and overlap-ratio metrics were also computed and show trends consistent with MAE and MPIW; they are omitted here for brevity.

**TABLE 4 QRF Model Performance**

| Metric | Training | Testing |
|---|---|---|
|  |  |  |
| **Vehicle-level ($N$=1,846,088)** |  |  |
| PICP (50%) | 49.65% | 48.55% |
|  |  |  |
| **Window-level ($N$= 2,827)** |  |  |
| MPIW (mph) | 17.84 | 18.08 |
| MAE ($Q_{50}$) (mph) | 0.73 | 1.55 |
| Accuracy ($Q_{50}$) ($\pm 5$ mph) | 99.91% | 96.43% |

### *Generalization Analysis Across Weather Conditions*

Our QRF model generalizes well across clear, rainy, and snowy conditions, see in Table 5. The 50% prediction interval stays close to the ideal coverage, with vehicle-level PICP values of 49.09% in clear weather, 45.8% in rain, and 45.7% in snow. The MPIW is stable at approximately 18–19 mph, indicating consistent uncertainty quantification. Median-speed accuracy naturally declines as conditions worsen: within $\pm 5$ mph it drops from 97.35% in clear weather to 89.69% in rain and 88.24% in snow. However, widening the tolerance to $\pm 6$ mph raises accuracy to 93.44% in rain and 94.12% in snow, showing that most errors remain small and operationally acceptable. Overall, the stable interval widths and calibrated PICP confirm that the model delivers reliable speed recommendations under diverse weather scenarios.

**TABLE 5 Testing Dataset Performance by Weather Condition**

| Weather | # Window samples | Vehicle-level PICP (50%) | Window-level MPIW (mph) | MAE ($Q_{50}$)(mph) | Accuracy ($Q_{50}$) ($\pm$5 mph) | Accuracy ($Q_{50}$) ($\pm$6 mph) |
|---------|------------------|--------------------------|--------------------------|---------------------|----------------------------------|----------------------------------|
| Clear | 2,490 | 49.09% | 18.01 | 1.41 | 97.35% | 98.35% |
| Rain | 320 | 45.80% | 18.52 | 2.50 | 89.69% | 93.44% |
| snow | 17 | 45.70% | 19.04 | 3.52 | 88.24% | 94.12% |

*Compare with Baseline*

Posted ± 10 % is a simple rule that clamps speed to a fixed legal interval (e.g., 55 mph $\pm$10%: [49.5, 60.5] mph). On the testing set, only 34.19% of vehicle trajectories fall inside this band, far below the nominal 50 % target, showing that a $\pm$10% tolerance around the posted limit does not capture the central half of observed speeds.

Rolling IQR uses purely historical data: for each 10-minute window we compute $[Q_{25}, Q_{75}]$ and the median from the previous N windows. Shrinking the window length from 24 (4 h) to 6 (1 h) windows boosts performance, with accuracy rises from 94.9% to 97.24% and MAE drops from 1.63 mph to 1.18 mph, because a shorter history "memorises" the most recent traffic state. The trade-off is poorer robustness: short windows are highly sensitive to noise or sudden state changes and deteriorate when few past samples are available.

QRF learns conditional quantiles from weather and traffic features rather than raw history. It keeps PICP and MPIW comparable to the Rolling IQR, attains a competitive MAE of 1.55mph, and sustains 96.43% accuracy without relying on a short memory.

NGBoost was tuned with $n_{estimators} = 200$, learning rate 0.05, and a normal output distribution for fair comparison. It edges out QRF by only 0.23% in PICP and 0.01 mph in MPIW (differences that are practically negligible), but its point prediction is markedly worse: MAE is double that of QRF and ±5 mph accuracy is 8.32% lower. Thus, while NGBoost attains similar interval calibration, QRF remains the more accurate and robust model overall.

Overall, QRF offers the best balance between calibrated prediction intervals and precise point estimates. Table 6 summarizes the baseline comparison.

**TABLE 6 Baseline Comparison**

| Method | Vehicle-level PICP (50%) | Window-level MPIW (mph) | MAE ($Q_{50}$) (mph) | Accuracy ($Q_{50}$) ($\pm$5mph) | Notes |
|--------|--------------------------|--------------------------|----------------------|--------------------------------|-------|
| Posted $\pm$ 10% | 34.19% | 11.00 | 2.66 | 89.00% | Fixed band [49.5, 60.5] mph |
| Rolling IQR - 6 window (1h) | 49.88% | 18.20 | 1.18 | 97.24% | History-only, short memory |
| Rolling IQR -12 window (2h) | 49.85% | 18.28 | 1.42 | 96.14% | History-only, short memory |
| Rolling IQR - 24 window (4h) | 50.07% | 18.20 | 1.63 | 94.90% | History-only, short memory |
| NGBoost | 48.78% | 18.09 | 3.21 | 88.11% | Parametric, well-calibrated but noisy $Q_{50}$ |
| QRF | 48.55% | 18.08 | 1.55 | 96.43% | Non-parametric, balanced |

**Case Study**

To assess real-world applicability, the model was applied to every 55 mph freeway segment in the Buffalo area, including: Niagara Thruway (I-190), New York State Thruway (I-90), Youngmann Expressway (I-290), Kensington Expressway (NY-33), Buffalo Skyway (NY-5), and Aurora Expressway (NY-400).

Each 10-minute window aggregates all connected-vehicle records observed on this freeway network. We examined two long testing periods. Period A: 19[th] − 28[th] Oct 2022 (early-winter transition); Period B: 23[rd] − 31[st] Mar 2023 (late-winter thaw) (Figure 3).

There are four key findings as follows.

1. Consistent interval coverage − The predicted $Q_{25} − Q_{75}$ interval nearly overlaps the ground-truth band across both periods, producing a vehicle-level PICP within ±2 % of the 50 % target. This demonstrates the model's ability to capture the 50 % speed distribution across varied weather conditions.

2. Accurate median prediction − The predicted median tracks closely with the true median over time. Particularly in March 2023 (Period B), the two curves exhibit near-complete overlap. This visual coherence confirms the model's ability to generalize across unseen test windows with high precision.

3. Effective Weather Responsiveness − On specific dates like October 19[th], 2022, March 23[rd], and March 28[th], 2023, notable dips in both the predicted median and $v_{phys}$ coincide with poor weather conditions (e.g., low visibility, reduced grip). This suggests the model is successfully integrating weather-sensitive variables and reflecting their impact on safe driving behavior.

4. Safety-Aware Recommendations − By comparing predicted medians and $v_{phys}$ against the fixed statutory speed limit (55 mph), it's evident that the model suggests slower safe-speed intervals when environmental conditions deteriorate—without being overly conservative. This provides a strong case for context-aware, data-driven speed guidance that balances realism and safety.

Across two testing periods encompassing all 55 mph freeways in Buffalo, NY, the QRF + $v_{phys}$ approach consistently produces well-calibrated, accurate, and safety-aware speed intervals. These results confirm the model's ability to operate network-wide and adapt to rapidly changing weather conditions.

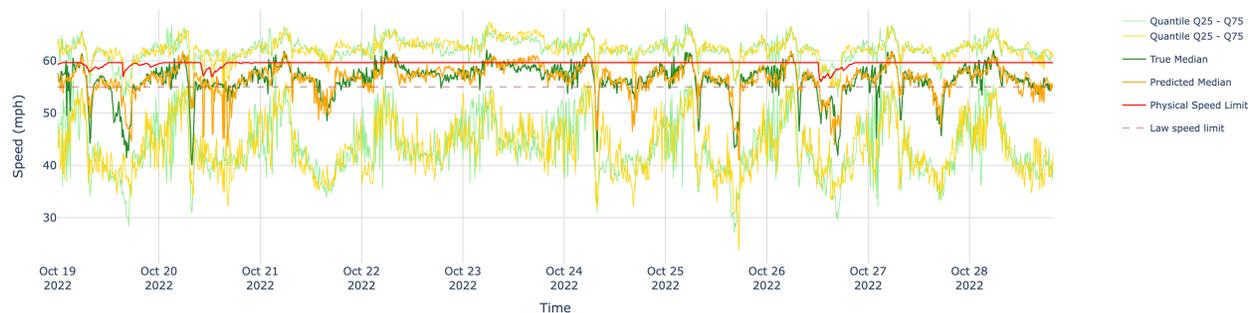

Figure 3a. Predicted and observed speed quantiles for all 55 mph freeways in Buffalo, NY, during 19[th] − 28[th] Oct 2022. The plot shows the predicted $Q_{25} − Q_{75}$ interval (light green), ground-truth $Q_{25} − Q_{75}$ interval (light yellow), predicted median $Q_{50}$ (orange), ground-truth

median (dark green), the physics-based upper limit $v_{\text{phys}}$ (solid red), and the statutory 55 mph limit (dashed line).

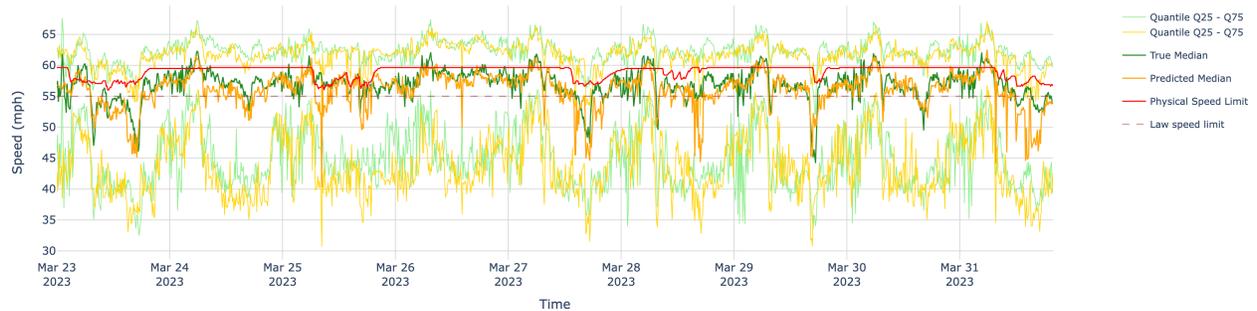

Figure 3b. Same as Fig. 3a for 23rd – 31st Mar 2023.

## CONCLUSIONS

This study presents a hybrid data-driven and physics-informed framework for real-time safe-speed interval recommendation on freeways, integrating CV data, RWIS data, and OSM road attributes. Focusing on 55 mph freeway segments in the Buffalo–Niagara metropolitan area, the approach leverages 73 days of high-resolution, multi-source data comprising over 6.6 million records.

The six key conclusions are as follows.

1. Quantile Regression Forests Provide Accurate and Calibrated Predictions — The proposed QRF model accurately estimates conditional speed quantiles in diverse weather conditions, achieving high prediction interval coverage probability (PICP = 48.55%), low mean absolute error (MAE = 1.55 mph), and 96.43% of predictions within $\pm 5$ mph. These results demonstrate robust generalization across unseen data and validate the model's reliability for real-world deployment.

2. Weather-Responsive Speed Estimation — The model effectively incorporates environmental factors such as pavement grip, visibility, and precipitation. Performance remains consistent under clear, rainy, and snowy conditions, with only modest declines in accuracy during inclement weather. The results confirm that the QRF model adapts to adverse conditions without excessive conservatism.

3. Physics-Based Constraints Enhance Safety Awareness — By integrating a physics-derived upper speed limit ($v_{phys}$) into the recommended interval, the method ensures that the suggested upper bound never exceeds the speed at which a vehicle can safely stop within the available sight distance. This safety-aware cap aligns well with drops in grip and visibility during poor weather, reflecting real-world constraints on vehicle operation.

4. Hybrid Interval Fusion Balances Data-Driven Insights and Physical Safety — The final recommended interval combines the QRF-predicted interquartile range with a posted legal limit ($v_{law}$) and the physics-based speed cap. This design produces intervals that are both empirically grounded and safety conscious. The method maintains tight interval widths while reducing the risk of unsafe speed recommendations during degraded road conditions.

5. Superior Performance Over Baseline Methods — Compared to static rules (e.g., posted speed $\pm 10\%$) and historical baselines (e.g., rolling IQR), the QRF approach offers better coverage, narrower intervals, and lower prediction errors. NGBoost performs comparably in terms of interval calibration but is less accurate in median prediction. Overall, QRF strikes the best balance between robustness, precision, and operational utility.

6. Network-Level Deployment Demonstrates Practical Value − The approach was successfully applied across all 55 mph freeway corridors in Buffalo, NY, over two separate testing periods. Results show consistent alignment between predicted and observed speed distributions, timely responsiveness to weather changes, and safe-speed interval recommendations aligned with physics constraints.

In summary, this study demonstrates the feasibility and effectiveness of a hybrid quantile–physical framework for freeway speed recommendation. By combining connected vehicle analytics with environmental sensing and physical principles, the proposed approach supports real-time, location-specific, and weather-aware speed guidance—advancing both traffic safety and intelligent transportation system capabilities.

## ACKNOWLEDGMENTS

The authors would like to thank the University at Buffalo for the partial funding of this research, and for the help and support from UB Information Technology and Facilities personnel.

## AUTHOR CONTRIBUTIONS

The authors confirm contribution to the paper as follows: study conception and design: Zhang, Qiao, and Sadek; data collection and processing: Zhang; analysis and interpretation of results: Zhang and Sadek; draft manuscript preparation: Zhang, Qiao and Sadek. All authors reviewed the results and approved the final version of the manuscript.